%% file: main.tex
\documentclass[10pt,twocolumn,letterpaper]{article}

\usepackage[pagenumbers]{cvpr} 

\input{preamble}

\definecolor{cvprblue}{rgb}{0.21,0.49,0.74}
\usepackage[pagebackref,breaklinks,colorlinks,citecolor=cvprblue]{hyperref}

\def\paperID{*****} 
\def\confName{CVPR}
\def\confYear{2024}

\title{Evaluating Zero-Shot GPT-4V Performance on 3D Visual \\Question Answering Benchmarks}

\author{Simranjit Singh\textsuperscript{1}, Georgios Pavlakos\textsuperscript{2}, Dimitrios Stamoulis\textsuperscript{1}\\
\textsuperscript{1} \textit{CoStrategist} R\&D Group, Microsoft Mixed Reality  \textsuperscript{2}University of Texas at Austin\\
}

\begin{document}

\twocolumn[{
\maketitle\centering
\captionsetup{type=figure}
\includegraphics[width=0.95\textwidth]{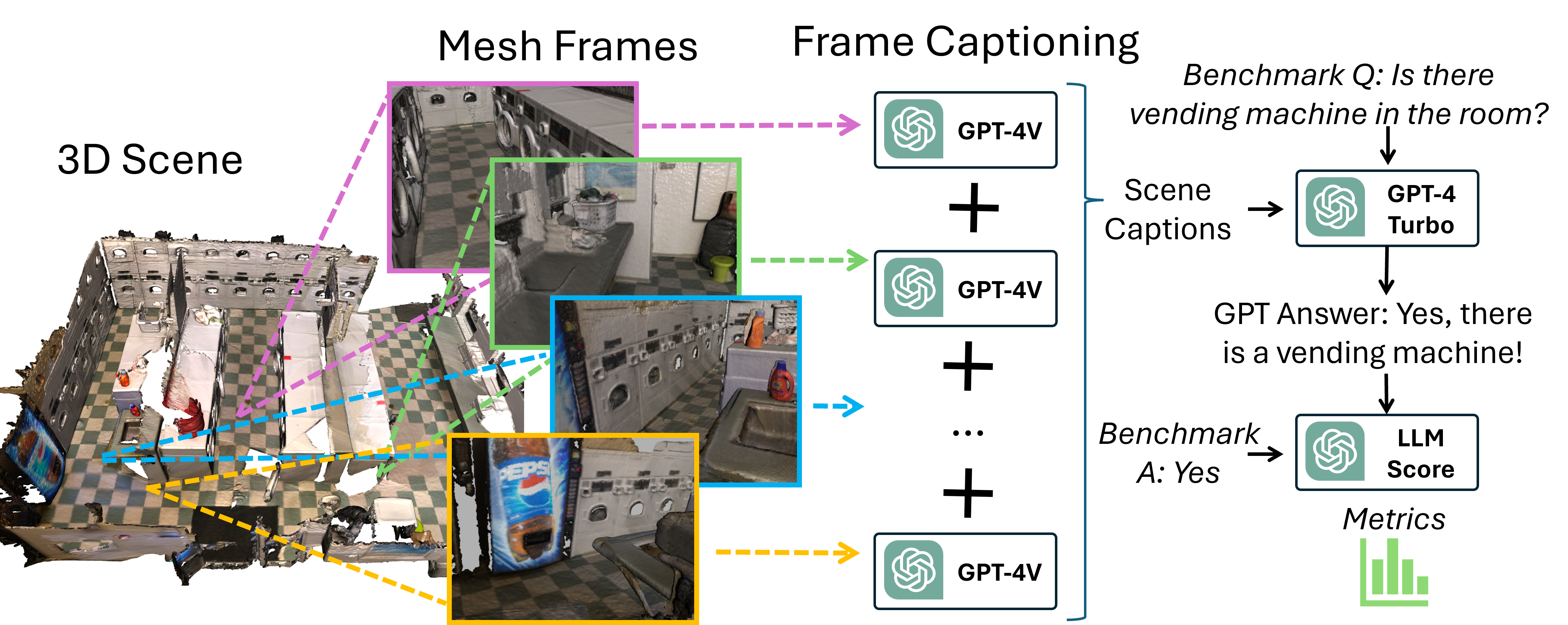}
\captionof{figure}{Illustration of a zero-shot GPT-4V agent answering 3D Visual Question Answering (VQA) questions. We present a preliminary investigation of how recently introduced open-vocabulary LLMs 
perform against older well-established closed-vocabulary benchmarks, namely \texttt{3D-VQA}~\cite{etesam20223dvqa} and \texttt{ScanQA}~\cite{azuma2022scanQA}. To stimulate future research and alleviate the extensive computational cost, we will be releasing all prompts, scene captions, GPT responses and results across all \texttt{ScanNet} scenes in our project repo: \textcolor{magenta}{\textit{Code available upon acceptance}}.}
\label{fig:agent}\vspace{5mm}
}]

\maketitle
\input{sec/0_abstract}    
\input{sec/1_intro}
\input{sec/2_relatedwork}
\input{sec/3_methodology}

\input{sec/5_experiments}
\input{sec/6_discussion}
\input{sec/7_conclusion}
{
    \small
    \bibliographystyle{ieeenat_fullname}
    \bibliography{main}
}
\end{document}

%% file: preamble.tex
\usepackage[dvipsnames]{xcolor}

\usepackage[accsupp]{axessibility}  
\usepackage{tcolorbox}
\usepackage{colortbl}

\usepackage{multirow}

%% file: sec/0_abstract.tex
\begin{abstract}
As interest in ``reformulating'' the 3D Visual Question Answering (VQA) problem in the context of foundation models grows, it is imperative to assess how these new paradigms influence existing closed-vocabulary datasets. In this case study, we evaluate the zero-shot performance of foundational models (GPT-4 Vision and GPT-4) on well-established 3D VQA benchmarks, namely 3D-VQA\cite{etesam20223dvqa} and ScanQA\cite{azuma2022scanQA}. We provide an investigation to contextualize the performance of GPT-based agents relative to traditional modeling approaches. We find that GPT-based agents without any finetuning perform on par with the closed vocabulary approaches. Our findings corroborate recent results that ``blind'' models establish a surprisingly strong baseline in closed-vocabulary settings. We demonstrate that agents benefit significantly from scene-specific vocabulary via in-context textual grounding. By presenting a preliminary comparison with previous baselines, we hope to inform the community’s ongoing efforts to refine multi-modal 3D benchmarks.
\end{abstract}

%% file: sec/1_intro.tex
\section{Introduction}
\label{sec:intro}

Advancements in 3D scene understanding have rendered AI agents essential for applications spanning from augmented reality to robotics~\cite{sermanet2023robovqa}. Yet, to make these technologies accessible to non-experts, it is critical for AI to interpret 3D scenes through natural language~\cite{OpenEQA2023}, where AI agents incorporate natural language understanding and ``common sense'' in sync with human perception and cognition~\cite{yang2023dawn}.

The fusion of vision and language has become a staple in multimodal 3D scene comprehension, with Visual Question Answering (VQA) tasks thoroughly examined across various benchmarks~\cite{azuma2022scanQA, achlioptas2020referit3d, etesam20223dvqa}. However, the rise of foundation models has posed new challenges, prompting a shift from closed- to open-vocabularies in VQA benchmarks and towards metrics that facilitate the assessment of open-ended responses. Punctuating this trend is the recently unveiled OpenEQA benchmark ~\cite{OpenEQA2023}.

In this paper, we present an exploratory analysis of how GPT-based agents perform against the well established 3D-VQA benchmarks, specifically \texttt{3D-VQA} and \texttt{ScanQA}. Our study reveals insights into the zero-shot capabilities of GPT-4V, providing a performance ``floor'' that helps contextualize agent performance relative to traditional, GPT-free baselines that are trained using extensive datasets. We demonstrate that zero-shot, finetuning-\textit{free} GPTs with almost off-the-shelf prompting, are remarkably competitive, achieving scores within 10\% of meticulously crafted DNN-based baselines. Moreover, our findings corroborate the surprising outcomes recently observed for open-vocabulary datasets within closed-set benchmarks: text-only or ``blind'' LLM agents -- responding to queries without visual data -- achieve surprisingly robust performance relying on language priors as ``common sense''~\cite{metaopenEQA}. Lastly, we discover that GPT-V, despite its free-form and open-ended nature, significantly benefits from the inclusion of scene-specific vocabulary during its captioning tasks.

To efficiently investigate agents  we implement a massively parallel task scheduler \cite{singh2024geoengine, singh2024evaluating} that facilitates concurrent VQA prompt execution across multiple GPT-4V \& GPT-4 Turbo endpoints. With the extensive computational effort already invested, we are releasing all prompts, scene captions, GPT responses and results over all \texttt{ScanNet} scenes in our project repo.\footnote{\textcolor{magenta}{\textit{Code available upon acceptance}}} We hope that this open resource stimulates research amidst the emerging discussion of rethinking 3D VQA for the era of foundation models.


%% file: sec/2_relatedwork.tex
\section{Related Work}
\label{sec:related_work}

\textbf{Existing ``closed-vocabulary'' VQA.}
Visual Question Answering (VQA) has been a key task across numerous AI and computer vision benchmarks thanks to its inherently multimodal nature, which involves the perception of both textual and visual input modalities. The importance of VQA is reflected in its application across a diverse range of domains, from embodied AI, situated reasoning, object localization, recognition, to activity detection, temporal window localization, and future forecasting~\cite{wu2021star, hou2023groundnlq, li2023seed, lei2018tvqa, das2018embodied, wijmans2019embodied, yu2019multi}. Given the multimodality, VQA is particularly vital for 3D scene understanding and analysis, with several well-established benchmarks building upon the ScanNet dataset such as 3D-VQA~\cite{etesam20223dvqa}, ScanQA~\cite{azuma2022scanQA}, and ReferIt3D~\cite{achlioptas2020referit3d}. However, these 3D VQA benchmarks were established before the advent of transformer-based open-vocabulary models, leading to their textual components being more closed-vocabulary.

\textbf{``open-vocabulary'' VQA.} In recent months, there is an emerging interest around developing new benchmarks that are better suited  to open-vocabulary models, such as MMBench~\cite{liu2023mmbench} and RoboVQA~\cite{sermanet2023robovqa}. Notably, Meta's release of the OpenEQA benchmark~\cite{OpenEQA2023} -- concurrent to our work -- punctuates this shift and provides a basis for researching multimodal embodied/3D agents. However, initial results are presented with respect to the newly introduced benchmark, lacking comparative analyses on previously established benchmarks. Our investigation aims to bridge this gap, providing insight into the performance of state-of-the-art agents when applied to 3D-VQA~\cite{etesam20223dvqa} and ScanQA~\cite{azuma2022scanQA} tasks. To our knowledge, our work is the first to conduct such analysis.

%% file: sec/3_methodology.tex
\section{Methodology}
\label{sec:approach}

In this section, we introduce the various agents, datasets, and metrics employed in our analysis. Our primary objective is to evaluate the \textit{zero-shot} capabilities of finetuning-\textit{free} GPT agents against established 3D VQA benchmarks.

\subsection{VQA Agents}

Our agent's workflow is illustrated in \cref{fig:agent}. Given a scene RGB-D sensor stream, we uniformly sample \textit{mesh} frames. These frames provides different views of a 3D scene mesh and we control the sampling by using frame sample rate parameter $F$ which means every $F$th frame is sampled from the stream. We then employ GPT-V as a powerful captioning model to produce per-frame captions that capture the detailed characteristics of different items within the scene. These captions are then provided as the \textit{language scene-description} to GPT-4 Turbo, which is tasked with responding to the benchmark questions.

While our methodology aligns closely with state-of-the-art 3D agent pipelines as outlined in~\cite{OpenEQA2023}, we make two deliberate implementation choices to tailor our approach to the benchmarks under consideration. First, for a more representative comparison with \texttt{3D-VQA} baselines, we employ \textit{mesh} images as inputs to our captioning model, rather than ``real'' 2D images.
Second, we utilize GPT-V to \textit{directly} caption the scene frames. An alternative would be utilizing dedicated captioning models, such as LLaVA-1.5~\cite{liu2023llava} or ConceptGraphs~\cite{gu2023conceptgraphs}, to create \textit{scene-graph} representations that incorporate 3D object semantics. However, OpenEQA \cite{OpenEQA2023} shows that 3D descriptions do not significantly impact agent performance, hence we postulate that the format of semantic representation can be more flexibly determined and generated directly via GPT-V prompting.

Given our overall goal to implement a \textit{fully} GPT-driven approach, we instruct GPT-V to craft Scene-Graph Captions-like (SGC) representations directly. Through in-context instructions and prompting, GPT-V is guided to ``encode'' the various aspects of the scene—detailing objects and their attributes such as color, size, and relative position - into a coherent textual scene-graph description. We explore two distinct captioning schemes. In a purely open-vocabulary framework, GPT-V is prompted to \textit{freely} describe the scene without any predefined information:

\begin{tcolorbox}[title=Prompting GPT-V for Vocabulary-agnostic SGC, colback=gray!20, colframe=gray!75, rounded corners, sharp corners=northeast, sharp corners=southwest]
\footnotesize
\texttt{\textbf{Prompt:} This is a view of 3D mesh from an indoor scene. List out all the objects you can see in the image. For each object provide color, shape, location and neighbouring objects following the object-description format [..]}
\end{tcolorbox}

Furthermore, we investigate a vocabulary \textit{grounded} approach, where GPT-V is given additional context. For this, we extract the scene-specific list of objects from the benchmark ground truths and we instruct GPT-V to concentrate on these specified items during its description process: 

\begin{tcolorbox}[title=Prompting GPT-V for Vocabulary-grounded SGC, colback=gray!20, colframe=gray!75, rounded corners, sharp corners=northeast, sharp corners=southwest]
\footnotesize
\texttt{\textbf{Prompt:} This is a view of 3D mesh from an indoor scene. List out all the objects you can see in the image based on the following items:\\
\textbf{Scene Items}: \{...\}\\
For each object provide color, shape, location and neighbouring objects following [..]}
\end{tcolorbox}


\subsection{Agents Types}

To answer all the benchmark questions, we prompt GPT-4 with the following prompt, providing some in-context examples from the train set. We ask multiple questions in a single API call and define the batch size of questions as $Q$

\begin{tcolorbox}[title=GPT-4 Turbo Prompting, colback=gray!20, colframe=gray!75, rounded corners, sharp corners=northeast, sharp corners=southwest]
\footnotesize
\texttt{You are provided with descriptions of various views of a 3D indoor scene. Your task is to answer questions about objects, their count, their attributes, and how they relate to other objects [..] Try your best to provide the answer and please format your answer like some of the QA templates provided below.\\
~\\
\textbf{In-context examples}: \{train-set QA pairs\}\\
\textbf{Scene Captions:} \{GPT-V captions\}\\
\textbf{User Query}: \{question\}\\
}
\end{tcolorbox}

By varying the \texttt{\textbf{Scene Captions}} input to GPT-4, we consider the following agents:

\colorbox{RoyalBlue}{\textbf{Blind.}} The agent answers ``blindly'' based only on the question and \textbf{without} any visual information, \textit{i.e.}, without any captions. This ``reference'' agent serves as a lower performance based on purely prior world knowledge and/or random guesses~\cite{OpenEQA2023}.

\colorbox{Goldenrod}{\textbf{Socratic w/ Vocabulary-agnostic SGC.}} The agent answers based on the scene description generated by GPT-V \textbf{without} scene-specific vocabulary.

\colorbox{BurntOrange}{\textbf{Socratic w/ Vocabulary-grounded SGC.}} The agent answers based on the scene description after providing GPT-V with the scene-specific \texttt{\textbf{Scene Items}} vocabulary.


\subsection{Benchmarks}

We conduct our analysis on two well-established 3D VQA benchmarks, namely \texttt{ScanQA} \cite{azuma2022scanQA} and \texttt{3D-VQA} \cite{etesam20223dvqa}. Both benchmarks are based on the \texttt{ScanNet} dataset~\cite{dai2017scannet}, which comprises scenes captured in various indoor environments, including bedrooms and offices. To facilitate a representative comparison with previously published baselines, we report metrics across all 71 scenes from the \texttt{ScanNet} validation set. The agents are evaluated on all the 53,395 and 4,675 questions from \texttt{3D-VQA} and \texttt{ScanQA}, respectively.

\subsection{Metrics}

To accurately compare with \texttt{ScanQA} and \texttt{3D-VQA} baselines, we evaluate the agent's answers using all the metrics reported in the respective papers: for \texttt{ScanQA}, we compute exact matching scores (EM@1, EM@10), BLEU scores (BLEU-1, BLEU-2, BLEU-3, BLEU-4), ROUGE-L, METEOR, CIDEr, and SPICE scores. Since our agent provides only a single answer, the metrics EM@1 and EM@10 are same in our evaluation. For \texttt{3D-VQA}, we report both the overall accuracy and accuracy per question type (i.e., counting, location, query-attribute, Yes/No).

Furthermore, consistent with observations from~\cite{OpenEQA2023}, we note that the open-vocabulary nature of VQA tasks presents challenges when evaluated using exact matching-based metrics or BLEU/ROUGE scores, which compare similarity versus a set of possible closed-vocabulary answers. Following this insight, we confirm the relevance of using an LLM to evaluate the correctness of agent responses, which has been shown to more closely correlate with human-like answers~\cite{OpenEQA2023}. For comprehensive analysis, we directly employ the recently introduced LLM-based score (termed as LLM-Match) directly using its prompting with GPT-4 Turbo:

\begin{table*}[t]
\centering
  \resizebox{\textwidth}{!}{%
  \begin{tabular}{@{}lccccccccccc@{}}
  \toprule
    Model                & EM@1 & EM@10 & BLEU-1 & BLEU-2 & BLEU-3 & BLEU-4 & \textbf{ROUGE-L} & METEOR & CIDEr & LLM-Match~\cite{OpenEQA2023} \\
    \midrule
    \textbf{Results from}~\cite{azuma2022scanQA} @\textit{ScanQA-val}\\
    RandomImage+MCAN (real) & 19.19 & 48.15 & 23.71 & 15.41 & 11.81 & 0.00 & 28.90 & 10.92 & 53.83 & -- \\
    RandomImage+MCAN (mesh) & 18.59 & 47.81 & 22.12 & 14.49 & 11.72 & 7.69 & 27.61 & 10.32 & 50.93  & -- \\
    TopDownImage+MCAN & 12.71 & 41.50 & 14.82 & 8.21 & 16.54 & 0.74 & 19.33 & 7.57 & 33.14 & -- \\
    VoteNet+MCAN 	 &  17.33 & 45.54 & 28.09 & 16.72 & 10.75 & 6.24 & 29.84 & 11.41 & 54.68  & -- \\
    ScanRefer+MCAN (real) & 14.37 & 44.12 & 17.02 & 10.17 & 15.77 & 0.72 & 22.02 & 8.45 & 38.73 & --  \\
    ScanRefer+MCAN (mesh) & 14.57 & 43.27 & 16.71 & 9.71 & 13.62 & 0.64 & 21.82 & 8.32 & 38.35 & --  \\
    ScanRefer+MCAN (e2e) 	 &  18.59 & 46.76 & 26.93 & 16.59 & 11.59 & 7.87 & 30.03 & 11.52 & 55.41& --  \\
    ScanQA (single) &  {20.28} & {50.01} & {29.47} & {19.84} & {14.65} & {9.55} & {32.37} & {12.60} & {61.66}  & -- \\
    ScanQA (multiple) &  \textbf{21.05} & \textbf{51.23} & \textbf{30.24} & \textbf{20.40} & \textbf{15.11} & \textbf{10.08} & 33.33 & 13.14 & \textbf{64.86} & --  \\
    \midrule
    \textbf{GPT Agents} @\textit{ScanQA-val}\\
    \colorbox{RoyalBlue}{Blind}  & 14.59 & 14.59 & 28.70 & 13.55 & 6.78 & 3.81 & 30.92 & 13.54  & 53.59 & 37.6 \\
    \colorbox{Goldenrod}{Socratic Vocab-agnostic SGC} & 10.05 & 10.05 & 16.28 & 7.00 & 3.22 & 0.98 & 20.03 & 8.75  & 34.22 & 25.7 \\
    \colorbox{BurntOrange}{Socratic Vocab-grounded SGC} & 18.01 & 18.01 & 24.49 & 10.68 & 4.64 & 1.63 & \textbf{33.43} & \textbf{14.23} & 58.32 & 42.0 \\
    \midrule
    \colorbox{Orchid}{\textbf{Human}} @\textit{ScanQA-test}~\cite{azuma2022scanQA} & 51.6 & -- & -- & -- & -- & -- & -- & -- & -- \\
    \bottomrule
    \end{tabular}
    }
    \caption{Performance assessment on \texttt{ScanQA}-val (71 scenes) compared to previously published benchmark baselines and ablations. For each agent, we compute all metrics as in the original paper, as well as the recently introduced LLM-based scoring metric.}
    \label{table:scanqa}
\end{table*}

\begin{tcolorbox}[title=LLM-Score Prompting - As proposed in~\cite{OpenEQA2023}, colback=gray!20, colframe=gray!75, rounded corners, sharp corners=northeast, sharp corners=southwest]
\footnotesize
\texttt{You are an AI assistant who will help me to evaluate the response given the question, the correct answer, and extra answers that are also correct. To mark a response, you should output a single integer between 1 and 5 (including 1, 5). 5 means that the response perfectly matches the answer or any of the extra answers. 1 means that the response is completely different from the answer and all of the extra answers [..]}
\end{tcolorbox}

Based on the GPT-4 evaluator scores $\sigma$ over $N$ questions, we calculate the aggregate metric LLM-Match~\cite{OpenEQA2023}:
\begin{equation}
  C = \frac{1}{N} \sum_{i} \frac{\sigma_i - 1}{4} \times 100\%
  \label{eq:llm-score}
\end{equation}

%% file: sec/5_experiments.tex
\section{Experiments}
\label{sec:experiments}

\subsection{Experimental Setup}

We utilize the \texttt{ScanNet toolkit}~\cite{gitScanNet} to derive camera poses, based on which we extract captures from the high-resolution mesh per scene. We use frame sample rate of $F$ which means every $F$th frame is used to answer the question. The default value of $F$ is 50 unless specified and we explore ablations of this parameter in Section \ref{sec:ablation}. In order to caption all frames across all scenes efficiently, we have developed a massively parallel scheduler that assigns GPT-V captioning tasks concurrently across 50 GPT-V Azure OpenAI endpoints. We then aggregate these per-frame captions to form a comprehensive scene description, which serves as input for GPT-4 to answer benchmark questions. We set the temperature for both GPT-4 Turbo and GPT-V to 0.2.

To manage the question-answering process across benchmarks efficiently and handle the sheer volume of queries, we also deploy the scheduler to simultaneously launch GPT-4 queries across 100 GPT-4 Azure OpenAI endpoints. This approach not only ensures efficient ablations, but also helps us circumvent the Requests-Per-Minute (RPM) limits set by GPT APIs~\cite{azrOAI}. In addition to this, we ask $Q$ questions in a single API call to reduce the time to answer the entire dataset. We carry out ablations on the this batch size $Q$ in Section \ref{sec:ablation}. Thanks to this level of parallelization, we are able to complete a run of the 5,000-question ScanQA dataset in just one hour.

\subsection{\texttt{ScanQA} results}

Table \ref{table:scanqa} presents the comparative analysis of agent performances on \texttt{ScanQA}. The best vocabulary-grounded GPT agent, with $Q=1$, performs slightly better than the best ScanQA baseline for the ROUGE-L score, a remarkably competitive result for a finetuning-\textit{free}, zero-shot method utilizing almost off-the-shelf prompting. Although the exact matching metrics (EM@ and BLEU-) fall below the top-performing ScanQA baseline, the ROUGE-L and METEOR scores are comparable or even superior. This discrepancy is expected, as the baselines are finely tuned through extensive training on specific datasets, which biases them toward the format of the final answer. In contrast, our GPT-based agents are deployed directly without such tailored training.
These observations partially support the discussion regarding the limitations of conventional similarity scores~\cite{OpenEQA2023}. It is interesting to note, however, that there is a discernible degree of correlation between ROUGE-L and LLM-Match based on the LLM scores for each agent.

\textbf{Blind GPT.}
Adding to the conversation that OpenEQA initiated recently~\cite{OpenEQA2023} that ``\textit{language provides an easier prior about the world}''~\cite{metaopenEQA}, we confirm that bling agents exhibit unexpectedly strong performance on \texttt{ScanQA} as well. As we discuss in detail later, we observe that GPT can employ ``common sense'' to achieve apt guesses: for instance, GPT is likely to respond correctly to questions such as ``\textit{what is the material of the kitchen counter}.''

\textbf{Vocabulary-grounding helps with GPT-V captioning.} We note a significant drop in performance when GPT-V is tasked with open-ended, ungrounded scene descriptions. Qualitatively inspecting the vocabulary-agnostic captions across several scenes, we observe a degradation in the quality of scene descriptions in two ways: first, without specific prompts directing attention to certain items, GPT-V faces difficulty in consistently grounding objects within the frames, especially smaller objects that are positioned on larger pieces of furniture, such as decorations. Second, GPT tends to use more commonly preferred synonyms (e.g., "couch" instead of "divan"), leading to a mismatch with the underlying object ``label''. An intriguing direction for future work, which we are currently exploring, is to employ ``oracle'' 3D objects from ScanNet scenes (\textit{e.g.}, replacing Detic 3D bounding-boxes~\cite{zhou2022detecting} with ``gold'' detections from the ReferIt3D ScanNet-based dataset~\cite{achlioptas2020referit3d}) to ground the Sparse Voxel Maps, LLaVA, or GPT-V captions. Should this improve GPT-4 performance further, it would suggest that properly grounding the scene description substantially influences the agent's overall results.

\begin{table*}[t]
\centering
  \begin{tabular}{@{}lcccccc@{}}
  \toprule
    Model    &    Location & Query & Count &  Yes/No & Overall & LLM-Match~\cite{OpenEQA2023}\\
    \midrule
    \textbf{Results from}~\cite{etesam20223dvqa} @\textit{3DVQA-val/test}\\
    Random (Q-type) & 0.2 & 5.0 & 2.0 & 50.0 & 11.08 & -- \\
    Majority (Q-type) & 4.6 & 19.0 & 32.0 & 58.0 & 27.44 & -- \\
    LSTM &  --  & --  & --  & --  & 42.02 & -- \\
    LSTM + BEV(2D) &  5.0 & 47.0 & 46.2 & 82.2 & 42.52 & -- \\
    LSTM + VoteNet & 11.6 & 44.3 & 50.0 & 82.0 & 42.98 & --\\
    NSM (pred) &  12.4 & 44.7 & 47.5 & 75.0 & 37.12 & --\\
    \midrule
    \textbf{GPT Agents} @\textit{3DVQA-val (53.4k questions)} \\
    \colorbox{RoyalBlue}{Blind}  &  1.71 & 28.75 & 31.82 & 59.89 & 26.27 & --\\
    \colorbox{BurntOrange}{Socratic Vocab-grounded SGC} &  10.25 & 29.94 & 32.93 & 65.19 & 30.46 & --\\
    \midrule
    \textbf{GPT Agents} @\textit{3DVQA-\textbf{mini}val (4.7k questions)} \\
    \colorbox{BurntOrange}{Socratic Vocab-grounded SGC} & 10.32 & 31.13 & 34.80 & 65.30 &  31.40 & 40.9\\
    \midrule
    \colorbox{Orchid}{Human}~\cite{etesam20223dvqa}  & -- & -- & -- & \multicolumn{2}{c}{91.89} & --\\
    \bottomrule
    \end{tabular}
    \caption{Performance assessment on \texttt{3D-VQA}-val with previously published baselines and ablations. For each agent, we compute all metrics as in the original paper, as well as the recently introduced LLM-based scoring metric.}
    \label{table:3dvqa}
\end{table*}

\subsection{\texttt{3D-VQA} results}

Table \ref{table:3dvqa} presents the comparative analysis of agent performances on \texttt{3D-VQA}. The results indicate a more substantial gap than observed in \texttt{ScanQA}, with the best-performing agent trailing the top baseline by over 10\% in overall accuracy. This widened disparity arises partly because the model struggles significantly with counting questions. Given GPT-V's existing limitations in enumerating objects within even within ``simple'' COCO-style images, this challenge is exacerbated when extended to counting across the entire mesh. 

Interestingly, given OpenEQA showed that 3D camera pose information does not markedly improve GPT's performance in captioning, we speculate that accurately deduplicating object counts for comprehensive questions (\textit{e.g.}, ``\textit{How many chairs are in the room?}'') is a very challenging task for the agent. Indeed, a closer examination of the responses reveals that GPT tends to default to majority-based estimations - a behavior reflected by the fact that agent ``Counting'' scores are close to the noise-floor of the ``Majority Q'' (constant liar); in other words, GPT-4 tends to guess low-count responses (usually 1 or 2) which happen to be more probable in the benchmark.

Similarly, the ``blind'' GPT agent exhibits a relatively weaker baseline than in \texttt{ScanQA}, hinting that zero-shot performance might be significantly influenced by the question distribution within a given task. An insightful implication of this for the 3D VQA community might be: should OpenEQA had included a greater proportion of counting questions, the effectiveness of ``common sense'' predictions might be lower. 

Last, having observed first-hand the computational overhead or running tens of thousands of VQA queries against expensive LLMs, we inspect how the number of VQA tasks impacts performance. To this end, we created a ``minival'' \texttt{3D-VQA} set, by uniformly sampling from the original 53.4k question pool to a more manageable 4.7k, mirroring the size of \texttt{ScanQA}. We find that performance trends remained consistent, falling within a 2\% variance range~\cite{OpenEQA2023}. This suggests that merely expanding the number of questions, without a significant shift in the question distribution, may not offer substantially more insights with regards to agent assessment, corroborating our group's parallel findings on evaluating GPT Copilots across other domains~\cite{singh2024geoengine}.

%% file: sec/6_discussion.tex
\section{Ablations}


\label{sec:ablation}
In this section we provide a systematic analysis of the various experimental configurations impacting the performance of our best peforming agent - \texttt{Socratic Vocab-grounded SGC}, specifically tailored for the ScanQA benchmark. This decision allows for a more focused examination, given the substantial size of the 3DVQA benchmark, which encompasses over 53,395 questions.

Initially, we study the consistency of our GPT-based agents across different experimental runs. Then we explore the performance disparities between utilizing original RGB frames and rendered views from 3D meshes. This comparison bridges the gap in the existing literature, which predominantly focuses either on direct RGB data  \cite{OpenEQA2023} or reconstructed 3D models \cite{etesam20223dvqa, azuma2022scanQA}, thereby enhancing our understanding of how different visual inputs affect agent performance.

Subsequently, we analyze the influence of the number of frames used per question. This aspect highlights the trade-off between the computational demands (in terms of the number of calls to GPT-V) and the resultant accuracy. Reducing the number of frames per scene can decrease the number of GPTV calls, optimizing resource usage.

Lastly, we conduct ablations on the number of questions handled per API call. This study is aimed at discerning the balance between operational efficiency (in terms of token utilization) and model accuracy. By processing multiple questions in a single API request, the total number of API calls necessary can be reduced, thereby conserving tokens and enhancing overall efficiency.

\subsection{Output Consistency Across Multiple Runs}
In this section, we analyze the consistency of performance metrics across multiple runs under identical experimental conditions. This investigation aims to determine the degree of variation in the answers provided by our GPT-based agents when posed the same question repeatedly. For this study, we maintained a constant batch size of $Q=20$ and  frame sample rate of $F=50$. We also keep the same temperature setting across different runs, as outlined in Section \ref{sec:experiments}. Temperature, as a parameter in GPT models, controls the randomness of the output by adjusting the probability distribution over the generated tokens; a lower temperature generally results in more deterministic and focused outputs.

Table \ref{table:variance} presents the empirical results from these tests. Interestingly, the variance in performance metrics across multiple runs was minimal, indicating high consistency in the answers generated by the agent. This is further quantified by the reported mean and standard deviation of the metrics, where the standard deviation is notably low. Such consistency show the deterministic nature of the outputs, likely due to the lower temperature settings employed in our experiments. Additionally, the LLM-Match scores remained quite stable across different trials, emphasizing the robustness of the metric as well.


\begin{table*}[t]
\centering
  \resizebox{\textwidth}{!}{%
  \begin{tabular}{@{}cccccccccccc@{}}
  \toprule
    \textbf{Experiment Runs} & EM@1 & EM@10 & BLEU-1 & BLEU-2 & BLEU-3 & BLEU-4 & ROUGE-L & METEOR & CIDEr & LLM-Match~\cite{OpenEQA2023} \\
    \midrule
    Run 1 & 17.26 & 17.26  & 25.97 & 11.68 & 5.25 & 1.46 & 31.19 & 13.02  & 54.61  & 39.0 \\
    Run 2 & 17.11 & 17.11 & 25.92 & 11.93 & 5.50 & 2.21 & 30.57 & 12.73 & 54.25 & 38.8 \\
    Run 3 & 16.90 & 16.90 & 25.73 & 11.68 & 5.08 & 0.00 & 30.35 & 12.66 & 53.74 & 38.7 \\
    Run 4 & 17.33 & 17.33 & 25.43 & 11.49 & 5.08 & 2.04 & 30.49 & 12.77 & 53.92 & 38.4 \\
    \midrule 
    Mean & 17.15 & 17.15 & 25.76 & 11.69 & 5.22 & 1.42 & 30.65 & 12.79 & 54.13 & 38.72 \\
    Std. & 0.03 & 0.03 & 0.05 & 0.03 & 0.03 & 1.01 & 0.13 & 0.02 & 0.14 & 0.06 \\
    \bottomrule
    \end{tabular}
    }
    \caption{Agent output consistency across multiple runs on \texttt{ScanQA}-val (71 scenes). The variance across multiple runs for same experimental setting is low.}
    \label{table:variance}
\end{table*}

\subsection{Mesh vs. RGB Frames for VQA}
In this ablation study, we investigate the use of original RGB frames versus rendered views of 3D meshes for the visual question answering task, a departure from the prevalent approach of utilizing 3D models in previous works on these benchmarks. Quantifying the potential performance improvement achieved by using the original 3D RGB captures for foundational models can be useful in understanding the information lost during scene reconstruction and the utilization of 3D CNNs to answer questions. Leveraging original frames has the advantage of producing richer captions with more detailed information about objects in the scene, particularly smaller ones. Unlike many previous approaches that rely on deep learning networks, particularly 3D CNNs which limit the input to only 3D models of scenes, our method allows for a direct comparison between using original RGB frames and mesh frames just by using the desired image to generate GPT-V captions and then using these captions for answering questions by agents.

Table \ref{table:rgb} presents the comparison between RGB and mesh frames. For this experiment, we maintain a default batch size ($Q$) of 20 and frames per scene ($F$) of 50. We observe a substantial increase in all scores, particularly notable improvements in ROUGE-L, METEOR, and CIDEr, showing an improvement of 13.15\%, 12.6\% and 18.39\% respectively. Additionally, all BLEU scores exhibit significant increases as well. This observation supports the intuitive notion that original RGB frames contain more information compared to their rendered 3D counterparts, resulting in richer captions and better answers. RGB frames capture intricate details, colors, textures, and lighting conditions directly from the scene, providing a comprehensive representation of spatial and semantic information. This richness enables GPT-V to understand scene dynamics more accurately and generate more descriptive captions. Additionally, RGB frames preserve subtle nuances such as shadows, reflections, and object occlusions, contributing to the overall realism of the scene. Leveraging this additional information leads to improved performance in visual question answering tasks.

\begin{table*}[t]
\centering
  \resizebox{\textwidth}{!}{%
  \begin{tabular}{@{}cccccccccccc@{}}
  \toprule
    \textbf{Image Source} & EM@1 & EM@10 & BLEU-1 & BLEU-2 & BLEU-3 & BLEU-4 & ROUGE-L & METEOR & CIDEr & LLM-Match~\cite{OpenEQA2023} \\
    \midrule
    3D Renders & 17.26 & 17.26  & 25.97 & 11.68 & 5.25 & 1.46 & 31.19 & 13.02  & 54.61   & 39.0 \\
    RGB & 19.87 & 19.87 & 29.94 & 14.55 & 7.16 & 2.21 & 35.29 & 14.66 & 64.65 & 43.5 \\
    \bottomrule
    \end{tabular}
    }
    \caption{Performance assessment on \texttt{ScanQA}-val for different image sources - Original RGB captures vs Mesh frames.}
    \label{table:rgb}
\end{table*}

\subsection{Number of Frames per scene}
In this ablation study, we evaluate the impact of varying the number of frames used per scene on the performance of our visual question answering (VQA) system. For the VQA task, we uniformly sample every $F$-th frame from the RGB-D stream, with each frame being a 3D render of the mesh corresponding to an original RGB capture. Descriptive scene captions are generated for each selected frame using GPTV, and these captions are subsequently combined to provide contextual information for our GPT-based agents to answer the questions. Ideally, using a higher number of frames (a lower value of $F$) should result in more detailed and dense captions as certain objects may be missing in some frames, and having a greater number of frames compensates for these absences by providing additional details. However, this approach incurs higher costs due to two main factors: 1) the increased number of GPTV calls required to generate captions for more frames, and 2) the larger combined caption size that needs to be processed by the GPT agents, resulting in higher token consumption. The use of more tokens not only increases costs but also slows down the inference process, thereby extending the overall response time.

The results are presented in \ref{table:frames_per_scene}, where different values of $F$ were tested. A clear trend is observed: as the value of $F$ increases, leading to fewer frames being used, the accuracy metrics decrease significantly. This decrement is smooth and marked, with ROUGE-L scores dropping from 31.19 to 18.14, METEOR from 13.02 to 7.8, and CIDEr from 54.61 to 32.23, respectively. The highest scores are obtained when $F=50$, indicating that sampling every 50th frame provides an optimal balance of detail and computational efficiency. We stop at $F=50$, as further increasing $F$ becomes expensive, both in terms of monetary cost and time. Future studies could explore the potential of achieving the best possible scores by utilizing all available frames, thus providing a comprehensive understanding of the trade-offs between cost, speed, and accuracy in the context of frame sampling for VQA tasks.

\begin{table*}[t]
\centering
  \resizebox{\textwidth}{!}{%
  \begin{tabular}{@{}cccccccccccc@{}}
  \toprule
    \textbf{$F$} & EM@1 & EM@10 & BLEU-1 & BLEU-2 & BLEU-3 & BLEU-4 & ROUGE-L & METEOR & CIDEr & LLM-Match~\cite{OpenEQA2023} \\
    \midrule
    50 & 17.26 & 17.26  & 25.97 & 11.68 & 5.25 & 1.46 & 31.19 & 13.02  & 54.61   & 39.0 \\
    100 & 15.23 & 15.23 & 22.51 & 10.16 & 4.55 & 1.81 & 28.17 & 11.74 & 49.68 & 35.3 \\
    250 & 12.19 & 12.19 & 19.36 & 8.51 & 3.84 & 1.60 & 23.56 & 10.20 & 40.49 & 31.3 \\
    500 & 10.82 & 10.82 & 16.84 & 7.58 & 3.50 & 0.00 & 20.42 & 8.78 & 35.93 & 28.5 \\
    750 & 9.84 & 9.84 & 15.39 & 7.05 & 3.32 & 1.35 & 18.33 & 8.12 & 32.81 & 26.2 \\
    1000 & 9.63 & 9.63 & 14.88 & 6.64 & 3.17 & 1.27 & 18.14 & 7.80 & 32.23 & 25.8 \\
    \bottomrule
    \end{tabular}
    }
    \caption{Ablation of frame sample rate ($F$) on \texttt{ScanQA}-val (71 scenes). We use every $F$-th frame in the RGB-D stream to generate GPTV captions and use those to answer questions.}
    \label{table:frames_per_scene}
\end{table*}

\subsection{Batch size}
In this ablation, we examine the impact of varying the batch size $Q$, which represents the number of questions our agent addresses within a single API call. This investigation helps understand the trade-off between the consumption of tokens and the performance of the agent. By posing more questions per API call, the total number of API calls required decreases, which in turn reduces the total token usage. This efficiency gain stems from the significant size of the context provided to the agent, which consists of concatenated descriptions from all sampled frames. However, increasing the batch size also complicates the task at hand. Answering multiple questions simultaneously can be more challenging than responding to a single query, as the agent must maintain accuracy across diverse question contexts.

Table \ref{table:batchsize} encapsulates the results of this ablation study. It reveals a discernible trend where an increase in batch size correlates with a slight dip in the performance metrics of the agent. Specifically, as the batch size escalates from 1 to 20, the ROUGE-L score decreases from 33.43 to 30.18. Similarly, the METEOR and CIDEr scores diminish from 14.23 to 12.68 and from 58.32 to 53.41, respectively. This trend, with a few anomalies, corroborates the intuitive expectation that batching questions would slightly compromise performance. Nonetheless, the magnitude of the performance decrement is not very significant, suggesting that while there is a cost to batching in terms of accuracy, the reduction in token usage and API calls might justify this approach in scenarios where computational efficiency is prioritized.


\begin{table*}[t]
\centering
  \resizebox{\textwidth}{!}{%
  \begin{tabular}{@{}cccccccccccc@{}}
  \toprule
    \textbf{Batch Size} & EM@1 & EM@10 & BLEU-1 & BLEU-2 & BLEU-3 & BLEU-4 & ROUGE-L & METEOR & CIDEr & LLM-Match~\cite{OpenEQA2023} \\
    \midrule
    1 & 18.01 & 18.01 & 24.49 & 10.68 & 4.64 & 1.63 & 33.43 & 14.23 & 58.32 & 42.0 \\
    2 & 17.99 & 17.99 & 25.44 & 11.41 & 5.25 & 2.41 & 32.90 & 13.90 & 57.82 & 40.4 \\
    5 & 17.41 & 17.41 & 25.72 & 11.66 & 5.18 & 1.67 & 31.49 & 13.13 & 55.43 & 39.3 \\
    10 & 16.98 & 16.98 & 25.46 & 11.33 & 4.94 & 1.39 & 30.74 & 12.75 & 54.27 & 38.9 \\
    15 & 16.56 & 16.56 & 25.65 & 11.85 & 5.43 & 2.62 & 30.35 & 12.69 & 53.34 & 37.9 \\
    20 & 16.83 & 16.83 & 25.81 & 11.90 & 5.67 & 2.61 & 30.18 & 12.68 & 53.41 & 37.4 \\
    \bottomrule
    \end{tabular}
    }
    \caption{Ablation of batch size ($Q$) on \texttt{ScanQA}-val (71 scenes). We see slight performance degradation as number of questions in a single API call increase.}
    \label{table:batchsize}
\end{table*}

%% file: sec/7_conclusion.tex
\section{Conclusion \& Future Work}
\label{sec:conclusion}
In this study, we presented an initial investigation of how emerging GPT-based agents perform on existing 3D VQA benchmarks, namely \texttt{3D-VQA} and \texttt{ScanQA}. Our primary finding indicates that GPT agents, even when operating without visual data (``blind''), serve as competitive baselines, suggesting the effectiveness of ``common sense'' guessing. Furthermore, the integration of scene-specific vocabulary into GPT-V enhances its captioning performance.

We hope that our investigation stimulates further research on 3D VQA benchmarks. To this end, and especially given the considerable computational cost of such analysis, we will be releasing our entire workflow. Hopefully, we aid the ongoing discussion regarding the ``need to reformulate'' older closed-form benchmarks in the ``era of foundation models.'' With that, we would like to highlight some particularly exciting and promising areas for future work that we have identified through our research and that we are actively investigating.

\textbf{What are the ScanQA and 3D-VQA upper bounds for human agents and multi-frame GPT?} With the ability of LLM-based metrics to evaluate open-vocabulary answers, it becomes compelling to revisit human-in-the-loop evaluation for ScanQA and 3D-VQA benchmarks — areas not extensively covered in the original papers, where in fact the reported human performance was about focus was aligning model replies with closed-vocabulary human answers. Moreover, we note that the current GPT-V API supports multi-frame input, a feature we plan to incorporate into our ongoing analysis as in~\cite{OpenEQA2023}.


\textbf{To what extent does elaborate prompting improve agent performance?} Early  methodologies, including ours and the OpenEQA~\cite{OpenEQA2023} agents, utilize straightforward Chain-of-Thought prompts that essentially guide the model to ``Think step-by-step.'' Investigating the effect of advanced prompting strategies, like few-shot ReAct and Tree-of-Thought~\cite{yao2023react, yao2023tree, fore2024geckopt}, will capture whether these augment or even inadvertently hurt overall agent effectiveness.


\textbf{Can we overcome Force-A-Guess failures more effectively?} As captured by OpenEQA and confirmed in our analysis, Socratic LLMs often abstain from answering. Instead of defaulting to best-guess responses, we drawing inspiration from \textit{self-consistency} prompting~\cite{wang2023selfconsistency}: our initial investigation shows that repeatedly posing the same question achieves a reduction of these failures by up to threefold after four to five repeated queries! We are currently pursuing this further towards reducing Force-A-Guess failures.

\textbf{Could in-context visual grounding improve agent performance?} Our version of vocabulary grounding could be viewed as in-context textual grounding, which in turn prompts the question of whether similar visual grounding strategies could be employed directly at the GPT-V level. For example, recent work shows that grounding with coordinates or markers was beneficial in zero-shot GPT-V performance for segmentation tasks~\cite{yang2023setofmark}. Thus, we are intrigued to explore analogous techniques for 3D mesh. In fact, we can draw inspiration from the original ScanQA paper and its \texttt{TopDownImage} ablation: would presenting the overall top-down ``floorplan'' scene as a visual prompt to GPT-V aid in addressing deduplication/counting issues?